\documentclass[11pt]{article}
\usepackage{eacl2017}
\usepackage{times}
\usepackage{url}
\usepackage{latexsym}

\usepackage{tikz}
\usetikzlibrary{shapes,arrows}
\usepackage{graphicx}
\usepackage{caption}
\usepackage{subcaption}
\usepackage{amsmath}
\usepackage{CJK}

\usepackage[margins]{trackchanges} 
\addeditor{chuan}

\eaclfinalcopy 
\def\namecite{\newcite}
\newcommand\blfootnote[1]{%
  \begingroup
  \renewcommand\thefootnote{}\footnote{#1}%
  \addtocounter{footnote}{-1}%
  \endgroup
}

\title{Addressing the Data Sparsity Issue in Neural AMR Parsing}

\author{Xiaochang Peng$^{*1}$, Chuan Wang$^{*2}$, Daniel Gildea$^1$ \and Nianwen Xue$^2$\\
  $^1$University of Rochester\\ 
  \{xpeng, gildea\}@cs.rochester.edu \\
  $^2$Brandeis University\\ 
  \{cwang24, xuen\}@brandeis.edu
  }

\date{}

\begin{document}
\maketitle
\begin{abstract}
Neural attention models have achieved great success in different NLP tasks.\blfootnote{*Both authors contribute equally.} However, they have not fulfilled their promise on the AMR parsing task due to 
the data sparsity issue. In this paper, we describe a sequence-to-sequence model for AMR parsing and present different ways to tackle the data sparsity problem. We show that our methods achieve significant improvement over a baseline neural attention model and our results are also competitive against state-of-the-art systems that do not use extra linguistic resources.
\end{abstract}

\section{Introduction}
\begin{figure}
\begin{center}
\scalebox{0.50}{
    \includegraphics{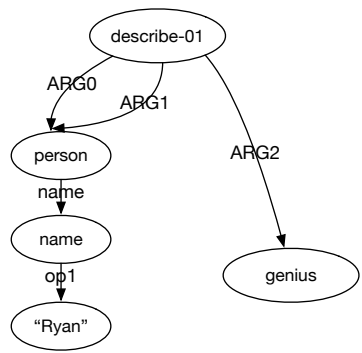}
}
\caption{An example of AMR graph representing the meaning of: ``Ryan's description of himself: a genius."}
\label{fig:amr-example}
\vspace{-1em}
\end{center}
\end{figure}

Abstract Meaning Representation (AMR) \cite{banarescu2013abstract} is a semantic formalism where the meaning
of a sentence is encoded as a rooted, directed graph. 
Figure~\ref{fig:amr-example} shows an example of an AMR in which the nodes represent the AMR concepts and the edges represent the relations between the concepts they connect.  
AMR concepts consist of predicate senses, named entity annotations, and in some cases, simply lemmas of English words. AMR relations consist of core semantic roles drawn from the Propbank \cite{palmer2005proposition} as well as very fine-grained semantic relations defined specifically for AMR.
These properties render 
the AMR representation useful in applications like question answering and semantics-based machine translation.

The task of AMR graph parsing 
is to map natural language strings to AMR semantic graphs.
Recently, a sizable new corpus of English/AMR pairs (LDC2015E86) has been released.  
Different parsers have been developed to tackle this problem~\cite{flanigan-EtAl:2014:P14-1,wang-xue-pradhan:2015:NAACL-HLT,artzi-lee-zettlemoyer:2015:EMNLP,pust-EtAl:2015:EMNLP,peng-song-gildea:2015:CoNLL}. 
Most of these parsers have used external resources such as
dependency parses, semantic lexicons, etc., to tackle the sparsity issue.

Recently, \namecite{sutskever2014sequence} introduced a neural network model for solving the general sequence-to-sequence problem, and \namecite{bahdanau2014neural} 
proposed a related model with an attention mechanism that is capable of handling long sequences. Both models achieve state-of-the-art results on large scale 
machine translation tasks.

However, sequence-to-sequence models mostly work well for large scale parallel data, usually involving millions of sentence pairs. 
\namecite{vinyals2015grammar} present a method which linearizes
parse trees into a sequence structure and therefore a sequence-to-sequence
model can be applied to the constituent parsing task. Competitive results have been achieved with an attention model on the Penn Treebank dataset, with only 40K annotated sentences.

AMR parsing is a much harder task in that the target vocabulary size is much larger, while the size of the dataset is much smaller. While for constituent parsing we only need to predict non-terminal labels and the output vocabulary is 
limited to 128 symbols, AMR parsing has both concepts and relation labels, 
and the target vocabulary size consists of tens of thousands of symbols.
\namecite{barzdins2016riga} applied a similar approach where AMR graphs are linearized using depth-first search and both concepts and relations are treated as tokens (see Figure~\ref{fig:amr-linearization}). 
Due to the data sparsity issue, their AMR parsing results are significantly lower than state-of-the-art models when using the neural attention model.

In this paper, we present a method which linearizes AMR graphs in a way that captures the interaction of concepts and relations.
To overcome the data sparsity issue for the target vocabulary, we propose a categorization strategy which first maps low frequency concepts and entity subgraphs to a reduced set of category types. In order to map each type
to its corresponding target side concepts, we use heuristic alignments to connect source side spans and target side 
concepts or subgraphs. During decoding, we use the mapping dictionary learned from the training data or heuristic rules for certain types to map
the target types to their corresponding translation as a postprocessing procedure.

Experiments show that our linearization strategy and categorization method are effective for the AMR parsing task. Our model improves significantly in comparison with 
the previously reported sequence-to-sequence results and provides a competitive benchmark in comparison with state-of-the-art results without using 
dependency parses or other external semantic 
resources.
\section{Sequence-to-sequence Parsing Model}
Our model is based on an existing sequence-to-sequence parsing model~\cite{vinyals2015grammar}, which is similar to models used in neural machine translation. 

\subsection{Encoder-Decoder}
\textbf{Encoder.} The encoder learns a context-aware representation for each position of the input sequence by mapping the inputs
$w_1,\dots,w_m$ into a sequence of hidden layers $h_1, \dots, h_m$. To model the left and right contexts of each input position, we use a bidirectional LSTM \cite{bahdanau2014neural}. First, each input's word embedding representation $x_1,\dots,x_m$ is obtained though a lookup table. Then these embeddings serve as the input to two RNNs: a forward RNN and a backward RNN. The forward RNN can be seen as a recurrent function defined as follows:
\begin{equation}
h_{i}^{fw} = f(x_i, h_{i-1}^{fw})
\end{equation}
Here the recurrent function $f$ we use is Long-Short-Term-Memory~(LSTM)~\cite{hochreiter1997long}. The backward RNN works similarly by repeating the process in reverse order. The outputs of forward RNN and backward RNN are then depth-concatenated to get the final representation of the input sequence.
\begin{equation}
h_{i} = [h_{i}^{fw}, h_{m-i+1}^{bw}]
\end{equation}
\textbf{Decoder.} The decoder is also an LSTM model which generates the hidden layers recurrently. Additionally, it utilizes an attention mechanism to put a 
``focus'' on the input sequence. At each output time step $j$, the attention vector $d_{j}^{'}$ is defined as a weighted sum of the input hidden layers, where the masking weight $\alpha_{i}^{j}$ is calculated using a feedforward neural network. Formally, the attention vector is defined as follows:
\begin{align}
u_{i}^{j} &= v^{T}\tanh(W_{1}h_{i}+W_{2}d_{j}) \\
\alpha_{i}^{j} &= \mathrm{softmax}(u_{i}^{j}) \\
d_{j}^{'} &= \sum_{i=1}^{m}\alpha_{i}^{j}h_{i}
\end{align}
where $d_{j}$ is the output hidden layer at time step $j$,
and $v$, $W_{1}$, and $W_{2}$ are parameters for the model. Here the weight vector $\alpha_{1}^{j},\dots,\alpha_{m}^{j}$ is also interpreted  as a soft alignment in the neural machine translation model, which similarly could also be treated as a soft alignment between token sequences and AMR relation/concept sequences in 
the AMR parsing task.
Finally, we concatenate the hidden layer $d_{j}$ and attention vector $d_{j}^{'}$ to get the new hidden layer, which is used to predict the output sequence label.
\begin{equation}
P(y_{j}|w, y_{1:j-1}) = \mathrm{softmax}(W_{3}[d_{j},d_{j}^{'}])
\end{equation}

\subsection{Parse Tree as Target Sequence}
\begin{figure}
\begin{center}
\scalebox{0.44}{
    \includegraphics{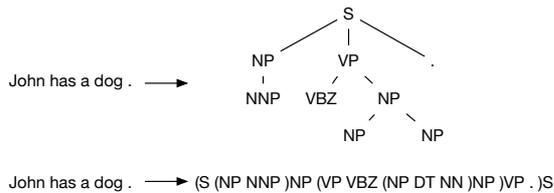}
}
\caption{Example parsing task and its linearization.}
\label{fig:parse-linearization}
\vspace{-1em}
\end{center}
\end{figure}
\namecite{vinyals2015grammar} designed a reversible way of converting the parse tree into a sequence, which they call linearization.
The linearization is performed in the depth-first traversal order.
Figure~\ref{fig:parse-linearization} shows an example of the linearization result. The target vocabulary consists of 128 symbols.

In practice, they found that using the attention model is more data efficient and works well on the parsing task.
They also reversed the input sentence and normalized the part-of-speech tags. After decoding, the output parse tree is recovered from the 
output sequence of the decoder in a post-processing procedure. Overall, the sequence-to-sequence model is able to match the performance of the Berkeley Parser~\cite{petrov-EtAl:2006:COLACL}.
\section{AMR Linearization}
\label{amrlinear}
\begin{figure}
\begin{center}
\scalebox{0.42}{
    \includegraphics{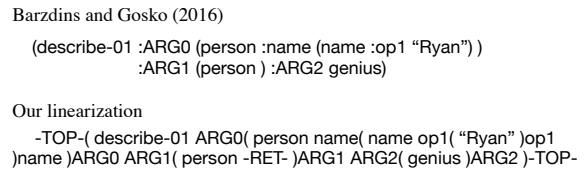}
}
\caption{Comparison of AMR linearization}
\label{fig:amr-linearization}
\vspace{-1em}
\end{center}
\end{figure}

\namecite{barzdins2016riga} present a similar linearization procedure where the depth-first traversal result of an AMR graph is used as 
the AMR sequence (see Figure~\ref{fig:amr-linearization}). The bracketing structure of AMR is hard to
maintain because the prediction of relation (with left parenthesis) and the prediction of an isolated right 
parenthesis are not correlated. As a result, the output AMR sequences usually have parentheses that do not match.

We present a linearization strategy which captures the bracketing structure of AMR and the connection between
relations and concepts. Figure~\ref{fig:amr-linearization} shows the linearization result of the AMR graph shown in Figure~\ref{fig:amr-example}. 
Each relation connects the head concept to a subgraph structure rooted at the tail concept, which shows one branch below the head concept.
We use the relation label and left parenthesis to show the beginning of the branch (subgraph) and use right parenthesis paired with
the relation label to show the end of the branch. We additionally add ``-TOP-(" at the beginning to show the start of the traversal of the AMR graph and add ``)-TOP-" at the
end to show the end of traversal. When a symbol is revisited, we replace the symbol with ``-RET-". We additionally add the revisited symbol before ``-RET-'' to decide where the reentrancy is introduced to.\footnote{This is an approximation because one concept can appear multiple times, and we simply attach the reentrancy to the most recent appearance of the concept. An additional index would be needed to identify the accurate place of reentrancy.}
We also get rid of variables and only keep the full concept label. For example, ``g / genius" to ``genius".

We can easily recover the original AMR graph from its linearized sequence. 
The sequence also captures the branching information of
each relation explicitly by representing it with a start symbol and an end symbol specific to that relation.
During our experiments, most of the output sequences have a matching bracketing structure using this linearization strategy. The idea of linearization is basically a depth-first traversal of the AMR where the original graph structure can be reconstructed with the linearization result. Even though we call it a sequence, its core idea is actually generating a graph structure from top-down.

\section{Dealing with the Data Sparsity Issue}
While sequence-to-sequence models can be successfully applied to constituent parsing, 
they do not work well on the AMR parsing task as shown by \namecite{barzdins2016riga}.
The main bottleneck is that the size of target vocabulary for AMR parsing is much larger than constituent parsing, tens of thousands in comparison with 128, and the size of training data is less than half of that available for parsing.

In this section, we present a categorization method which significantly reduces
 the target vocabulary size,
 as the alignment from the attention model does not work well on the relatively small dataset. To adjust for the alignment errors made by the attention model, we propose to add supervision from an alignment produced by an external aligner which can use lexical information to overcome the limit of data size.
\subsection{AMR Categorization}
\label{categorize}
\begin{figure*}
  \includegraphics[width=\textwidth,height=7cm]{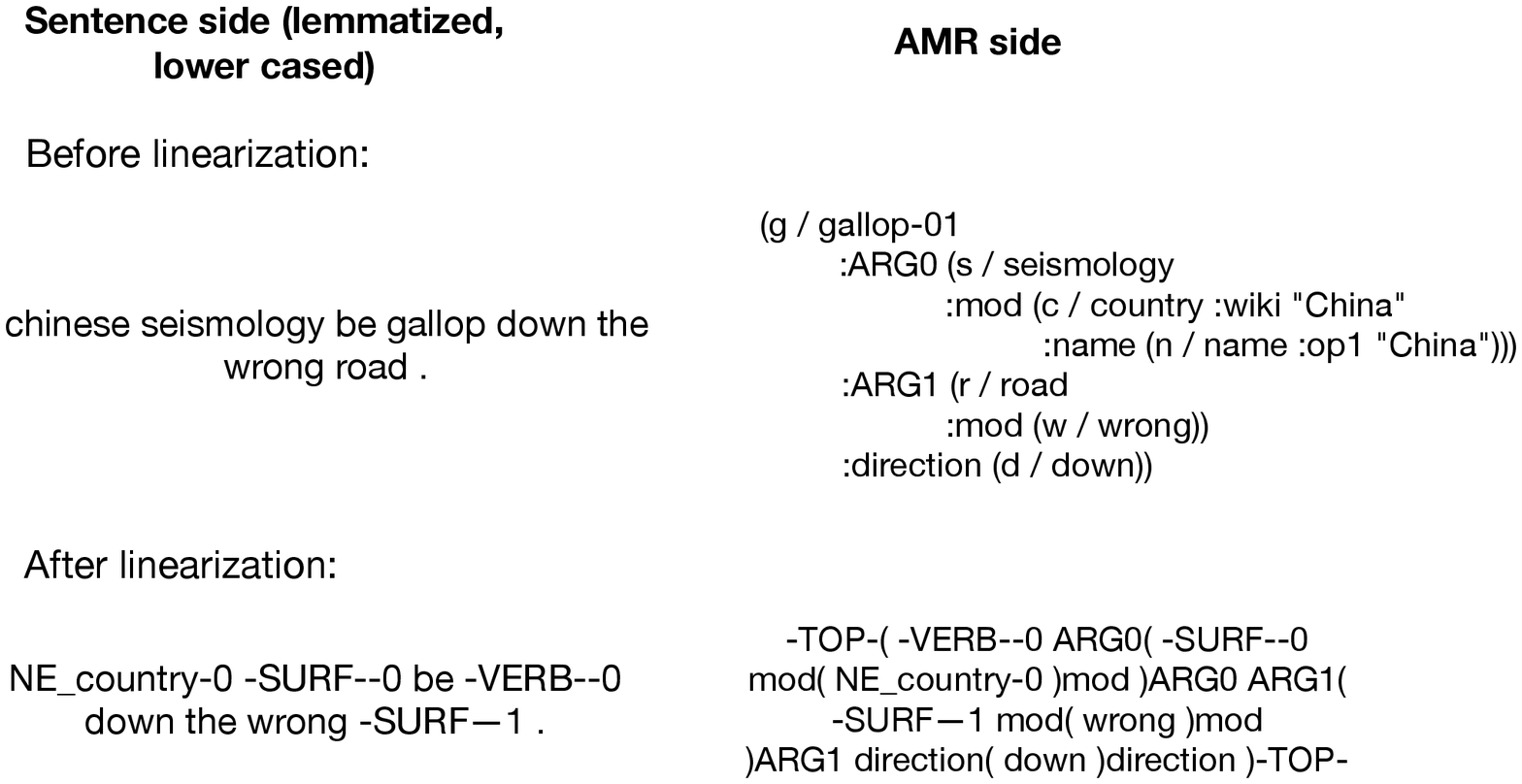}
  \caption{An example of categorized sentence-AMR pair.}
  \label{fig:categorize}
\end{figure*}

We define several types of categories and map low frequency words into these categories. 
\begin{enumerate}
\item DATE: we reduce all the date entity subgraphs to this category, ignoring details of the specific date entity. 
\item NE\_\{{\em ent}\}: we reduce all named entity subgraphs to this category, where {\em ent} is the root label of each subgraph, such as {\em country} or {\em person}.
\item -VERB-: we map predicate variables with low frequency ($n < 50$) to this category
\item -SURF-: we map non-predicate variables with low frequency ($n < 50$) to this category
\item -CONST-: we map constants other than numbers, ``-", ``interrogative", ``expressive'', ``imperative" to this category.
\item -RET-: we map all revisited concepts to this category.
\item -VERBAL-: we additionally use the verbalization list~\footnote{http://amr.isi.edu/download/lists/verbalization-list- v1.06.txt} from the AMR website and map matched subgraphs to this category.
\end{enumerate}
After the re-categorization, the vocabulary size is substantially reduced to around 2000, though this vocabulary size is still very large for the relatively small dataset.
These categories and the frequent concepts amount to more than 90\% of all the target words, and each of these are learned with a larger number of occurrences.
\subsection{Categorize Source Sequence}

The source side tokens also have sparsity issues. For example, even if we have mapped the number {\em 1997} to ``DATE", we can not easily generalize it to the token {\em 1993} if it does not appear in the training data. Also, some special 6-digit date formats such as ``YYMMDD" are 
hard to learn using co-occurrence statistics.

Our basic approach to dealing with this issue is to generalize these sparse tokens or spans to some special categories (currently we use the same set of categories defined in the previous section).
On the training data, we can use the heuristic alignment. For example, if we learned from the heuristic alignment that ``010911" is aligned
to a date-entity of September 11, 2001 on the AMR side, we use the same category ``DATE'' to replace this token. We distinguish this alignment from 
other date alignments by assigning a unique indexed category ``DATE-X" to both sides of the alignment, where ``X" counts from 0 and adds one 
for each new date entity from left to right on the sentence side. The same index strategy goes for all the other categories. Figure~\ref{fig:categorize} shows an example of the linearized parallel sequence. The first infrequent non-predicate variable ``seismology" is mapped to ``-SURF--0", then ``wrong'' to ``-SURF--1" based on its position on the sentence side. The indexed category labels are then projected onto the
target side based on the heuristic alignment. During this re-categorization procedure, we build a map $Q$ from each token or span to its most likely concept or category on the target side based on relative frequency. We also dump a DATE template for recognizing new date entities by abstracting away specific date fields such as ``1997'' to ``YEAR", ``September" to ``MONTH". For example, we build a template ``MONTH DAY, YEAR" from the specific date mention ``June 6, 2007".

During decoding, we are only given the sentence. We first use the date templates learned from the training data to recognize dates in each sentence. We also use a named entity tagger to recognize named entity mentions in the sentence. We use the entity name and wiki information from $Q$ if there is a match of the entity mention, otherwise for convenience we simply use ``person" as the entity name and use wiki ``-". For each of the other tokens, we first look it up in $Q$ 
and replace it with the most likely mapping. If there is no match, we further look it up in the verbalization list. In case there is still no match, we use the part of speech information to assign its category. We replace verbs with category ``-VERB-" and nouns with category ``-SURF-". In accordance with the categorized token sequence, we also index each category in the resulting sequence from left to right.
\subsection{Recovering AMR graph}
\label{recover}
During decoding, our output sequences usually have categories and we need to map each category to AMR concepts or subgraphs.
When we categorize the tokens in each sentence before decoding, we save the mapping from each category to its original token as table $D$. 
As we use the same set of categories on both source and target sides, we heuristically align target side category label to its source side counterpart from left to right.
Given table $D$, we know which source side token it comes from and use the most frequent concept or subgraph of the token to replace the category.
\subsection{Supervised Attention Model}
\label{superalign}
In this section, we propose to learn the attention vector in a supervised manner. There are two main motivations behind this. 
First, the neural attention model usually utilizes millions of data points to train the model, which learns a quite reasonable attention vector that at each output time step constrains the decoder to put a focus on the input sequences~\cite{bahdanau2014neural,vinyals2015grammar}. However, we only have 16k sentences in the AMR training data and our output vocabulary size is quite large, which makes it hard for the model to learn a useful alignment between the input sequence and AMR concepts/relations. Second, as argued by \namecite{2016arXiv160904186L}, 
the sequence-to-sequence model tries to calculate the attention vector and predict the current output label simultaneously. This makes it impossible for the learned soft alignment to combine information from the whole output sentence context. However, traditional word alignment can easily use the whole output sequence, which will produce a more informed alignment.

Similar to the method used by \namecite{2016arXiv160904186L}, we add an additional loss to the original objective function to model the disagreement between the reference alignment and the soft alignment produced by the attention mechanism. Formally, for each input/output sequence pair $(\mathbf{w},\mathbf{y})$ in the training set, the objective function is defined as:
\begin{equation}
-\frac{1}{n}\sum_{j=1}^{n}\log p(y_{j}|\mathbf{w}, y_{1:j-1}) + \lambda \Theta(\bar\alpha^{j}, \alpha^{j})
\end{equation}
where $\bar\alpha^{j}$ is the reference alignment for output position $j$, which is provided by the existing aligner, $\alpha^{j}$ is the soft alignment, $\Theta()$ is cross-entropy function, $n$ is the length of output sequence and $\lambda$ is the hyperparameter which serves as a trade-off between sequence prediction and alignment supervision. Note that the supervised attention part doesn't affect the decoder which will predict the output label given learned weights.

One issue with this method is how we represent $\bar{\alpha}$. As the output of conventional aligner is a hard decision, alignment is either one or zero for each input position. 
In addition, multiple input words could be aligned to one single concept. 
Finally, in the AMR sequences, there are many output labels (mostly relations) which don't align to any word in the input sentence. 
We utilize a heuristic method to process the reference alignment. We assign an equal probability among the words that are aligned to one AMR concept. Then if the output label doesn't align to any input word, we assign an even probability for every input word.

\section{Experiments}

We evaluate our system on the released dataset (LDC2015E86) for SemEval 2016 task 8 on meaning representation parsing~\cite{may:2016:SemEval}. The dataset contains 16,833 training, 1,368 development and 1,371 test sentences which mainly cover domains like newswire, discussion forum, etc.  All parsing results are measured by Smatch (version 2.0.2) \cite{cai-knight:2013:Short}. 

\subsection{Experiment Settings}
We first preprocess the input sentences with tokenization and lemmatization. Then we extract the named entities using the Illinois Named Entity Tagger~\cite{ratinov-roth:2009:CoNLL}. 

For training all the neural AMR parsing systems, we use 256 for both hidden layer size and word embedding size. Stochastic gradient descent is used to optimize the cross-entropy loss function and we set 
the drop out rate to be 0.5. We train our model for 150 epochs with initial learning rate of 0.5 and learning rate decay factor 0.95 if the model doesn't improve for the 3 last epochs.

\subsection{Baseline Model}
Our baseline model is a plain sequence-to-sequence model which has been used in the constituent parsing task~\cite{vinyals2015grammar}. While they use a 3-layer deep LSTM, we only use a single-layer LSTM for both encoder and decoder since our data is relatively small and empirical comparison shows that stacking more layers doesn't help in our case. AMR linearization follows Section \ref{amrlinear} without categorization. Since we don't restrict the input/output vocabulary in this case, our vocabulary size is quite large: 10,886 for output vocabulary and 2,2892 for input vocabulary. We set them to 10,000 and 20,000 respectively and replace the out of vocabulary words with \texttt{\_UNK\_}. 

\subsection{Impact of Re-Categorization}
We first inspect the influence of utilizing categorization on the input and output sequence. Table \ref{table:t1} shows the Smatch evaluation score on development set. 
\begin{table}[ht]
\begin{center}
\scalebox{0.93}{
\begin{tabular}{|l|c|c|c|}
\hline \bf System & \bf P & \bf R & \bf F \\ \hline
Baseline & 0.42 & 0.34 &  0.37\\
Re-Categorization ($n=50$) & 0.55 & 0.46 & 0.50 \\
\hline
\end{tabular}}
\end{center}
\caption{\label{table:t1}  Re-Categorization impact on development set}
\end{table}

We see from the table that re-categorization improves the F-score by 13 points on the development set. As mentioned in section \ref{categorize}, by setting the low frequency threshold $n$ to 50 and re-categorizing them into a reduced set of types, we now reduce the input/output vocabulary size to (2,000, 6,000). This greatly reduces the label sparsity and enables the neural attention model to learn a better representation on this small scale data. Another advantage of this method is that although AMR tries to abstract away surface forms and retain the semantic meaning structure of the sentence, a large portion of the concepts are coming from the surface form and have exactly same string form both in input sentence and AMR graph.  
For example, \texttt{nation} in sentence is mapped to concept \texttt{(n / nation)} in the AMR\@. For the frequent concepts in  the output sequence, since the model can observe many training instances, we assume that it can be predicted by the attention model. For the infrequent concepts, because of the categorization step, we only require the model to predict the concept type and its relative position in the graph. By applying the post-processing step mentioned in Section \ref{recover}, we can easily recover the categorized concepts to their original form.

\begin{figure}
\begin{center}
\scalebox{0.55}{
    \includegraphics{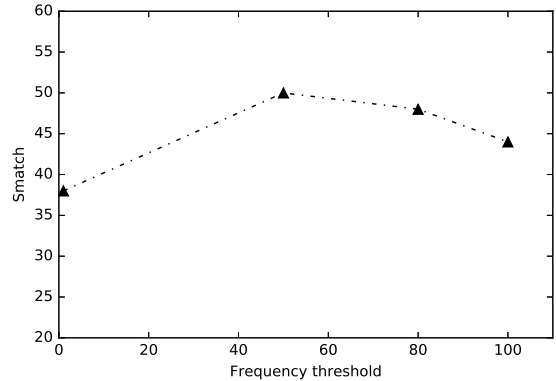}
}
\caption{AMR parsing performance on development set given different categorization frequency.}
\label{fig:amr-freq-tune}
\end{center}
\end{figure}

We also inspect how the value of re-categorization frequency threshold $n$ affects the AMR parsing result. As shown in Figure \ref{fig:amr-freq-tune}, setting $n$ to 0, which means no output labels will be categorized into types \textsc{-VERB-} and \textsc{-SURF-}, doesn't improve the baseline system. The reason is that we still have a large output vocabulary size and training data is still sparse with respect to the low frequency output labels. Also, if we set $n$ too high, although the output vocabulary size becomes smaller, some of the frequent output labels that the model handles well originally will be put into the coarse-grained types, losing information in the recovery process. Thus we can see from the plot that after the optimal point the Smatch score will drop. Therefore, we choose to set $n=50$ in the subsequent experiments.

\subsection{Impact of Supervised Alignment}
\textbf{Choice of External Aligner}.
There are two existing AMR aligners: one is a rule-based aligner coming with JAMR~\cite{flanigan-EtAl:2014:P14-1}, which defines regular expression patterns to greedily match between AMR graph fragment and input token spans; another one is an unsupervised aligner~\cite{pourdamghani-EtAl:2014:EMNLP2014} which adopts the traditional word alignment method in machine translation. 
Although evaluated on different set of manual alignment test sentences, both aligners achieved $\sim$90\% F1 score. Here we choose to use the second aligner, as it covers broader domains.
\vspace{1em}

\noindent\textbf{Different alignment configurations}
To balance between the sequence learning and alignment agreement, We empirically tune the hyperparameter $\lambda$ and set it to 0.3. For the external alignment we use for reference, we convert it to a vector with equal probability as discussed in Section \ref{superalign}. We then train a sequence-to-sequence model with re-categorized input/output and report the result on development set. 

\begin{table}[ht]
\begin{center}
\scalebox{0.95}{
\begin{tabular}{|l|c|c|c|}
\hline  System &  P &  R &  F \\ \hline
Baseline & 0.42 & 0.34 &  0.37\\
Categorization~($n=50$) & 0.55 & 0.46 & 0.50 \\
SuperAttn+Cate~($n=50$) & 0.56 & 0.49 & 0.52 \\
\hline
\end{tabular}}
\end{center}
\caption{\label{table:t2}  Supervised attention impact on development set}
\end{table}
As shown in Table \ref{table:t2}, the supervised attention model is able to further improve the Smatch score by another 2 points, which are mainly contributed by 3 points increase in recall. Since the reference/external alignment is mostly between the input tokens and AMR graph concepts, we believe that the supervised attention model is able to constrain the decoder so that it will output concepts which can be aligned to some tokens in the input sentence.

 \begin{table}[ht]
\begin{center}
\scalebox{0.95}{
\begin{tabular}{|l|c|c|c|}
\hline  System &  P &  R &  F \\ \hline
SuperAttn+Cate~($n=50$) & 0.56 & 0.49 & 0.52 \\
\textsc{No-Relation-Align} & 0.46 & 0.40 & 0.43 \\
\hline
\end{tabular}}
\end{center}
\caption{\label{table:t3}  Supervised attention impact on development set}
\end{table}

Because we have relations in the AMR graph, the alignment problem here is different from the word alignment in machine translation. To verify the effectiveness of our setup, we also compare our configuration to the condition \textbf{\textsc{No-Relation-Align}} where we only ignore the alignment between sentence and AMR relations by putting an all zero vector as the reference attention for each output relation label. From Table \ref{table:t3} we see that simply ignoring the reference attention for relations would greatly affect the model performance, and how we effectively represent the reference alignment for relations is crucial for the supervised attention model.

\subsection{Results}
In this section we report our final result on the test set of SemEval 2016 Task 8 and compare our model with other parsers. We train our model utilizing re-categorization and supervised attention with hyperparameters tuned on the development set. 
Then we apply our trained model on the test set.

Firstly, we compare our model to the existing sequence-to-sequence AMR parsing model of~\namecite{barzdins2016riga}. As shown in table \ref{table:t4}, the word-level model in~\namecite{barzdins2016riga} is basically our baseline model. The second model they use is a character-based sequence-to-sequence model. Our model can also be regarded as a word-level model; however, by utilizing carefully designed categorization and supervised attention, our system outperforms both their results by a large margin.
\begin{table}[ht]
\begin{center}
\scalebox{0.9}{
\begin{tabular}{|l|c|c|c|}
\hline  System &  P &  R &  F \\ \hline
\textbf{Our system} & \bf 0.55 & \bf 0.50 & \bf 0.52 \\
\namecite{barzdins2016riga}$^\dag$ & - & - & 0.37 \\ 
\namecite{barzdins2016riga}$^\star$ & - & - & 0.43 \\ 
\hline
\end{tabular}}
\end{center}
\caption{\label{table:t4}  Compare to other sequence-to-sequence AMR parser. \namecite{barzdins2016riga}$^\dag$ is the word-level neural AMR parser, \namecite{barzdins2016riga}$^\star$ is the character-level neural AMR parser.}
\end{table}

Table 5 gives the comparison of our system to some of the teams participating in SemEval16 Task 8. Since a large portion of the teams extend on the state-of-the-art system CAMR~\cite{wang-xue-pradhan:2015:NAACL-HLT,wang-xue-pradhan:2015:ACL-IJCNLP,wang-EtAl:2016:SemEval}, here we just pick typical teams that represent different approaches. We can see from the table that our system fails to outperform the state-of-the-art system. However, the best performing system CAMR uses a dependency structure as a starting point, where dependency parsing has achieved high accuracy recently and can be trained on larger corpora. Also, it utilizes semantic role labeling and complex features, which makes the training process a long pipeline. 
Our system only needs minimal preprocessing, and doesn't need the dependency parsing step. Our approach is competitive with the SHRG~(Synchronous Hyperedge Replacement Grammar) method of \namecite{peng-song-gildea:2015:CoNLL}, which does not require a dependency parser and uses SHRG to formalize the string-to-graph problem as a chart parsing task. However, they still need a concept identification stage, while our model can learn the concepts and relations jointly.
\begin{table}[ht]
\begin{center}
\begin{tabular}{|l|c|c|c|}
\hline  System &  P &  R &  F \\ \hline
Our system & 0.55 &  0.50 &  0.52 \\
\namecite{peng-gildea:2016:SemEval} & 0.56 & 0.55 & 0.55\\
CAMR & 0.70& 0.63 & 0.66 \\
\hline
\end{tabular}
\end{center}
\caption{\label{table:t5}  Comparison to other AMR parsers.}
\end{table}
\section{Discussion}
In this paper, we have proposed several methods to make the sequence-to-sequence model work competitively against conventional AMR parsing systems. Although we haven't outperformed state-of-the-art system using the conventional methods, our results show the effectiveness of our approaches to reduce the sparsity problem when training sequence-to-sequence model on a relatively small dataset. Our work could be aligned with the effort to handle low-resource data problems when building the end-to-end neural network model. 

In neural machine translation, the attention model is traditionally trained on millions of sentence pairs, while facing low-resource language pairs, the neural MT system performance tends to downgrade~\cite{2016arXiv160402201Z}. There has been growing interest in tackling sparsity/low-resource problem in neural MT\@. \namecite{2016arXiv160402201Z} use a transfer learning method to first pre-train the neural model on rich-resource language pairs and then import the learned parameters to continue training on low-resource language pairs so that the model can alleviate the sparsity problem through shared parameters. \namecite{firat-cho-bengio:2016:N16-1} builds a multilingual neural system where the attention mechanism can be shared between different language pairs. Our work could be seen as parallel efforts to handle the sparsity problem since we focus on the input/output categorization and external alignment, which are both handy for low-resource languages.

In this paper, we haven't used any syntactic parser. However, as shown in previous works~\cite{flanigan-EtAl:2014:P14-1,wang-xue-pradhan:2015:NAACL-HLT,artzi-lee-zettlemoyer:2015:EMNLP,pust-EtAl:2015:EMNLP}, using dependency features helps improve the
parsing performance significantly because of the linguistic similarity between the dependency tree and AMR structure. An interesting extension would be to
use a linearized dependency tree as the source sequence and apply sequence-to-sequence to generate the AMR graph. Our parser could also benefit from the modeling techniques in~\namecite{DBLP:journals/corr/WuSCLNMKCGMKSJL16}.

\section{Conclusion}
Neural attention models have achieved great success in different NLP tasks. However, they have not been as successful on AMR parsing due to 
the data sparsity issue. In this paper, we described a sequence-to-sequence model for AMR parsing and present different ways to tackle the data sparsity problems. We show that our methods have led to  significant improvement over a baseline neural attention model, and our model is also competitive against models that do not use extra linguistic resources.

\paragraph{Acknowledgments}  Funded in part by a Google Faculty Award.

\bibliography{eacl2017}
\bibliographystyle{eacl2017}
\end{document}